\documentclass[letterpaper, 10 pt, conference]{ieeeconf}  

\IEEEoverridecommandlockouts                              

\usepackage{graphics} 
\usepackage{epsfig} 
\usepackage{amsmath} 
\usepackage{amssymb}  
\usepackage{subfigure}
\usepackage{multirow}
\usepackage{etoolbox}
\makeatletter
\patchcmd{\@makecaption}
  {\scshape}
  {}
  {}
  {}
\makeatletter
\patchcmd{\@makecaption}
  {\\}
  {.\ }
  {}
  {}
\makeatother

\title{\LARGE \bf
Road Obstacle Detection based on Unknown Objectness Scores
}

\author{Chihiro Noguchi, Toshiaki Ohgushi, and  Masao Yamanaka
\thanks{Toyota Motor Corporation, Otemachi, Chiyoda-ku, Tokyo, 100-0004, Japan. Corresponding author is Chihiro Noguchi ({\tt\small chihiro\_noguchi\_aa@mail.toyota.co.jp})}%
}

\begin{document}

\maketitle
\thispagestyle{empty}
\pagestyle{empty}

\begin{abstract}
The detection of unknown traffic obstacles is vital to ensure safe autonomous driving.
The standard object-detection methods cannot identify unknown objects that are not included under predefined categories.
This is because object-detection methods are trained to assign a background label to pixels corresponding to the presence of unknown objects.
To address this problem, the pixel-wise anomaly-detection approach has attracted increased research attention.
Anomaly-detection techniques, such as uncertainty estimation and perceptual difference from reconstructed images, make it possible to identify pixels of unknown objects as out-of-distribution (OoD) samples.
However, when applied to images with many unknowns and complex components, such as driving scenes, these methods often exhibit unstable performance.
The purpose of this study is to achieve stable performance for detecting unknown objects by incorporating the object-detection fashions into the pixel-wise anomaly detection methods.
To achieve this goal, we adopt a semantic-segmentation network with a sigmoid head that simultaneously provides pixel-wise anomaly scores and objectness scores.
Our experimental results show that the objectness scores play an important role in improving the detection performance.
Based on these results, we propose a novel anomaly score by integrating these two scores, which we term as unknown objectness score.
Quantitative evaluations show that the proposed method outperforms state-of-the-art methods when applied to the publicly available datasets.

\end{abstract}

\section{INTRODUCTION}

\label{sec:intro}

When driving, various types of obstacles can be encountered on the road, and detecting them from in-vehicle camera images is essential for safe autonomous driving. 
However, standard object-detection methods \cite{redmon2018yolov3,NIPS2015_14bfa6bb,10.1007/978-3-030-58452-8_13} fail to detect unknown objects that do not belong to any predefined object categories.
It is also difficult to address this issue in an anomaly detection setting because unknown objects are learned to be identified as part of the image background.
Since many road obstacles cannot be classified as the predefined object categories, they are basically identified as part of the background.

Against this background, recently proposed methods \cite{Jung_2021_ICCV,Ohgushi_2020_ACCV,Di_Biase_2021_CVPR} are based on semantic-segmentation networks.
Semantic segmentation \cite{zhao2018icnet,10.1007/978-3-030-01234-2_49} aims to assign an appropriate label to almost every pixel, including the image background.
This makes it easier to consider anomaly detection settings for detecting unknown regions because both predefined object and background pixels are included in the in-distribution; unknown regions can be detected as out-of-distribution (OoD) samples.
This approach is known as pixel-wise anomaly detection \cite{Di_Biase_2021_CVPR}.
While the pixel-wise anomaly-detection methods are effective for detecting unknown regions in an image, their performance is often unstable when applied to road obstacle detection.
These methods do not limit detection to unknown objects, but can detect any unknown regions in the image.
In the context of the road obstacle detection, however, this also means an increase in false positive predictions, especially in the image background.

In this paper, we propose a novel road obstacle detection method that combines the principles of pixel-wise anomaly detection and object detection. Representative object-detection methods, such as Faster-RCNN \cite{NIPS2015_14bfa6bb} and YOLOv3 \cite{redmon2018yolov3}, calculate \textit{objectness scores} that measure how much the corresponding local region in an image is object-like. The use of objectness scores is promising for stabilizing the performance of detecting unknown objects, as is the case with detecting predefined objects. 
However, in fact, these objectness scores are not compatible with the anomaly-detection settings, as mentioned earlier. Therefore, we aim to address this issue by integrating the objectness scores with pixel-wise anomaly scores defined from predicted probabilities in semantic segmentation.

\begin{figure}[t!]
\centering
\includegraphics[width=8.7cm]{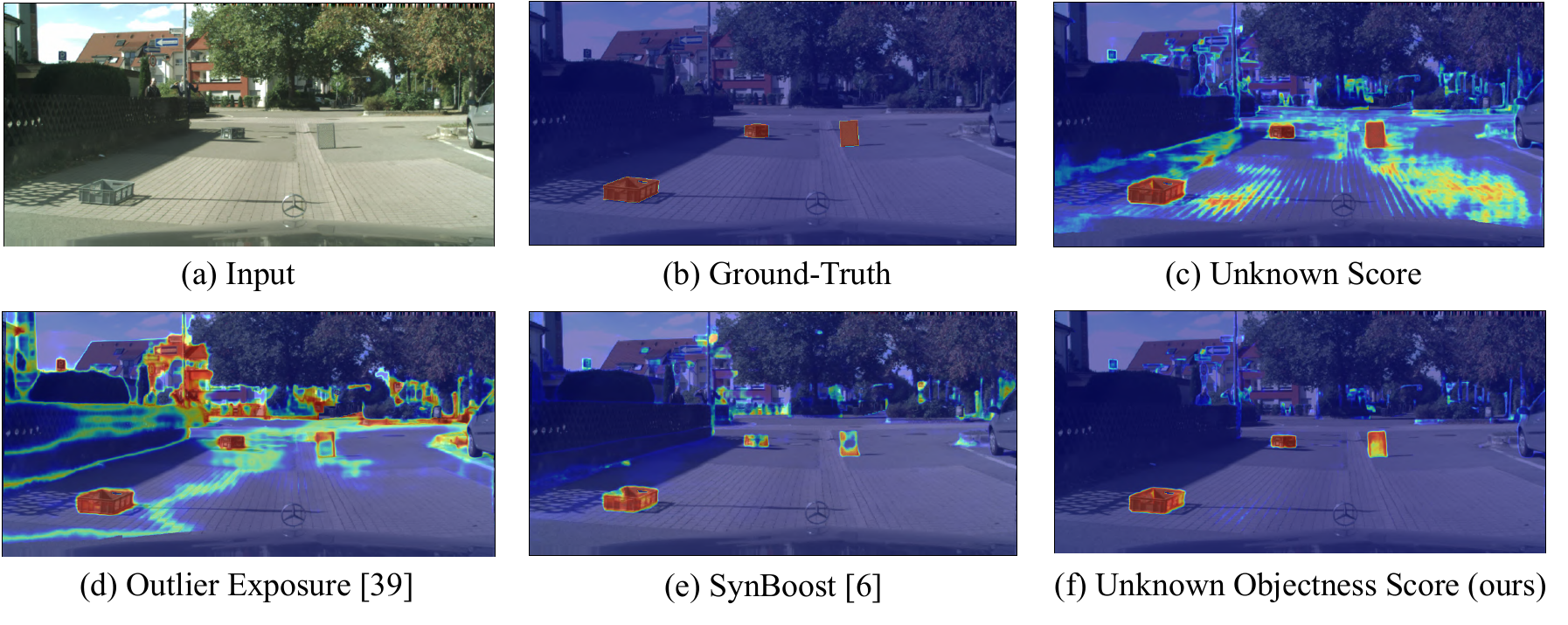}
\caption{Examples of road obstacle detection results using the proposed and representative methods. The proposed anomaly score---unknown objectness score---reduces the number of false-positive predictions, especially in the background regions, compared to the baseline scores (c) and the results of the existing methods (d,e).}
\label{fig:p1_fig}
\end{figure}

The proposed method is based on a semantic-segmentation network utilizing a sigmoid head. This approach allows for the assignment of multiple classes to each pixel, enabling a more flexible design of class labels compared to a typical softmax head. Specifically, we merge predefined object classes, such as cars and traffic signals, into an \textit{object class}, and define the objectness score as the predicted probability of the object class. The use of the sigmoid head allows the segmentation model to simultaneously learn the labels of both the object class and the predefined classes.
Additionally, we define \textit{unknown scores} as pixel-wise anomaly scores obtained from the outputs of the sigmoid head.
By combining the objectness and unknown scores, we define \textit{unknown objectness scores}.

This design is particularly effective when only a limited number of images of road obstacles are available.
In order to use these images for training our model, we assign the object class label to pixels corresponding to road obstacles.
This approach eliminates the need to create new classes for road obstacles and also precludes the necessity to limit the detection targets to specific objects.
This situation often arises in practical settings since encountering road obstacles in standard driving scenarios is infrequent.
We experimentally show the effectiveness of the unknown objectness scores using the publicly available datasets (Fig. \ref{fig:p1_fig}).

The major contributions of this study are as follows.
\begin{itemize}
    \item A novel road obstacle detection method is proposed based on a novel anomaly score---unknown objectness score---defined by the outputs obtained from the sigmoid head. 
    \item The proposed method exhibits better performance than state-of-the-art approaches, when applied to the publicly available datasets.
    \item The proposed method is based exclusively on a semantic-segmentation network; this facilitates fast detection without the need for additional modules.
\end{itemize}

\section{Related work}
\subsection{Pixel-wise Anomaly Detection}
Numerous pixel-wise anomaly detection methods have been proposed. These methods can be primarily classified into two approaches based on the concepts of autoencoder and uncertainty estimation.

\subsubsection{Autoencoder-based Approaches}
Autoencoder-based methods require the use of at least two modules---an encoder and a decoder.
The encoder maps input images into a feature space and the decoder subsequently attempts to reconstruct the original image.
If the test images contain an unknown region, the encoder and decoder cannot perform an accurate reconstruction. This facilitates the identification of the unknown region.
The encoder and decoder modules are, in general, configured using networks such as the generative adversarial network (GAN) \cite{goodfellow2014} and variational autoencoder (VAE) \cite{Kingma2014}.
Previous studies have proposed several methods based on different autoencoder modules, including the RBM autoencoder \cite{7225680}, VAE \cite{baur2019}, and GAN \cite{Schlegl2017,SCHLEGL201930,Zenati2018EfficientGA,8972475}.

Recent studies \cite{9010663,Xia2020,Ohgushi_2020_ACCV,Di_Biase_2021_CVPR} have proposed methods based on the adoption of a segmentation network as an encoder.
The decoder in these methods attempts to reconstruct an input image from a predicted semantic label map \cite{Chen_2017_ICCV,NEURIPS2019_b2eb7349}.
A segmentation network can fail to assign an accurate label map to an unknown region. This causes the subsequent decoder to generate a reconstructed image that features a large gap or an indeterminate spot in the unknown region.
To quantify the gap between the input and reconstructed images, these methods calculate the perceptual difference \cite{Chen_2017_ICCV}.
The perceptual difference is obtained by calculating the difference between both feature maps from the middle layers of the ImageNet \cite{ILSVRC15} pretrained VGG \cite{DBLP:journals/corr/SimonyanZ14a}.
Furthermore, some existing studies \cite{9010663,Di_Biase_2021_CVPR} introduced a discrepancy network and trained it to enhance this difference.

\subsubsection{Uncertainty-based Approaches}
Utilizing uncertainty estimation is another standard approach for performing pixel-wise anomaly detection.
The predicted semantic labels in an unknown region can be expected to demonstrate high uncertainty. This is because the semantic-segmentation model is trained using dataset that does not contain the ground truth of such regions.

Bayesian neural networks \cite{Gal2016Bayesian,pmlr-v48-gal16} based on Bernoulli’s approximation of the variational distribution are widely used for uncertainty estimation \cite{NIPS2017_2650d608,NIPS2017_9ef2ed4b}.
This approach, which is closely related to the dropout technique \cite{JMLR:v15:srivastava14a}, is referred to as the MC dropout.
Although the MC dropout is applicable to pixel-wise uncertainty estimation \cite{BMVC2017_57,huang2018efficient,7789580,mukhoti2018evaluating,Gustafsson_2020_CVPR_Workshops}, it encounters problems when applied to unknown object detection in images representing driving scenarios.
Specifically, the high-uncertainty pixels estimated by the MC dropout often do not match the positions of unknown objects, especially in complex situations such as driving \cite{9010663,Ohgushi_2020_ACCV}.
Another uncertainty-estimation approach involves the simple utilization of the softmax entropy \cite{hendrycks17baseline}.
The softmax entropy can be easily calculated from the predicted probabilities generated from a segmentation network.
To facilitate unknown object detection in driving scenarios, the use of the maximum logit instead of softmax entropy has been found to improve detection performance \cite{hendrycks2019anomalyseg}. In addition, a recent study \cite{Jung_2021_ICCV} has revealed that standardizing the maximum logit can further improve performance.

\begin{figure*}[t!]
\centering
\includegraphics[width=16cm]{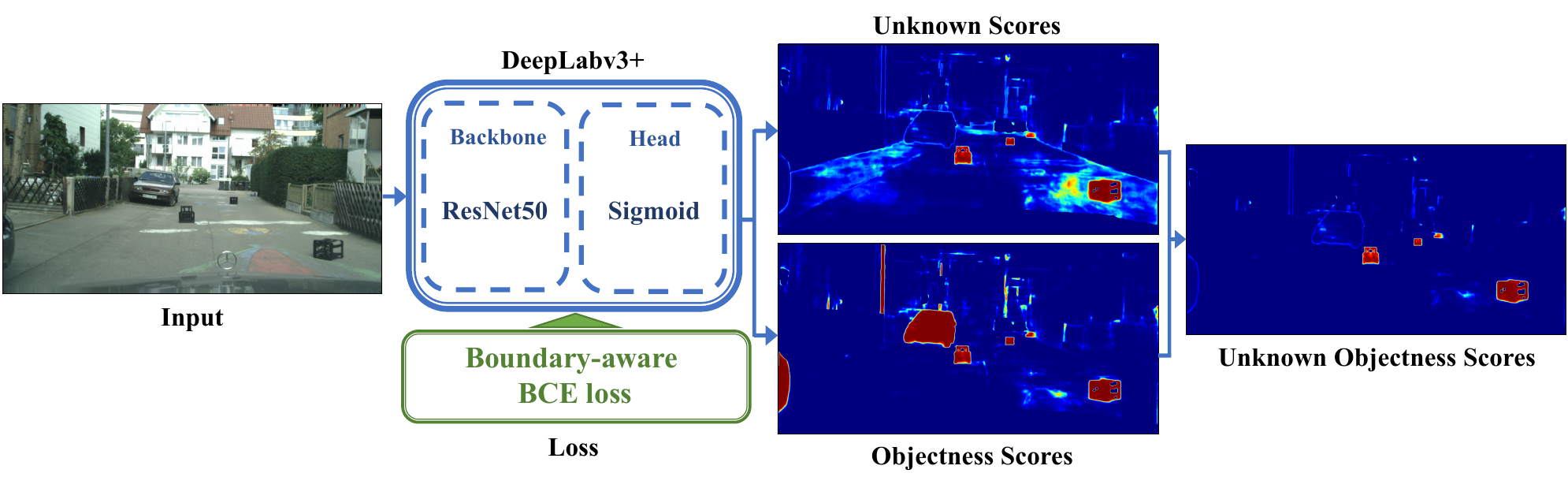}
\caption{Overview of the proposed method. The input image is fed to a semantic-segmentation network with a sigmoid head. Unknown objectness scores are obtained from the sigmoid outputs by combining unknown and objectness scores.}
\label{fig:overview}
\end{figure*}

\section{Proposed Method}
\label{sec:approach}
Our proposed method builds on a semantic-segmentation network with a sigmoid head.
This method leverages the properties of the sigmoid output to define the unknown objectness scores.
Fig. \ref{fig:overview} presents an overview of the proposed method.
An input image is first fed to the segmentation network, the output of which provides unknown scores and objectness scores (Sec. \ref{sec:one_vs_rest}). 
Then, the unknown objectness scores are defined by combining these two scores (Sec. \ref{sec:prior_knowledge}).
In addition, we introduce a loss function to reduce false positives in boundary regions (Sec. \ref{sec:loss_function}).

\subsection{Unknown Scores}
\label{sec:one_vs_rest}
State-of-the-art methods for performing pixel-wise anomaly detection in driving scenarios are often developed considering a semantic-segmentation network.
In these networks, the labeled data facilitate a pixel-level semantic understanding, which is essential for detecting unknown object pixels.
The proposed method is likewise built on a semantic-segmentation network. 
However, unlike most methods that adopt a softmax head in the output of the final layer, the proposed method adopts a sigmoid head.

\begin{figure}[t!]
\centering
\subfigure[]{
\includegraphics[width=4.cm]{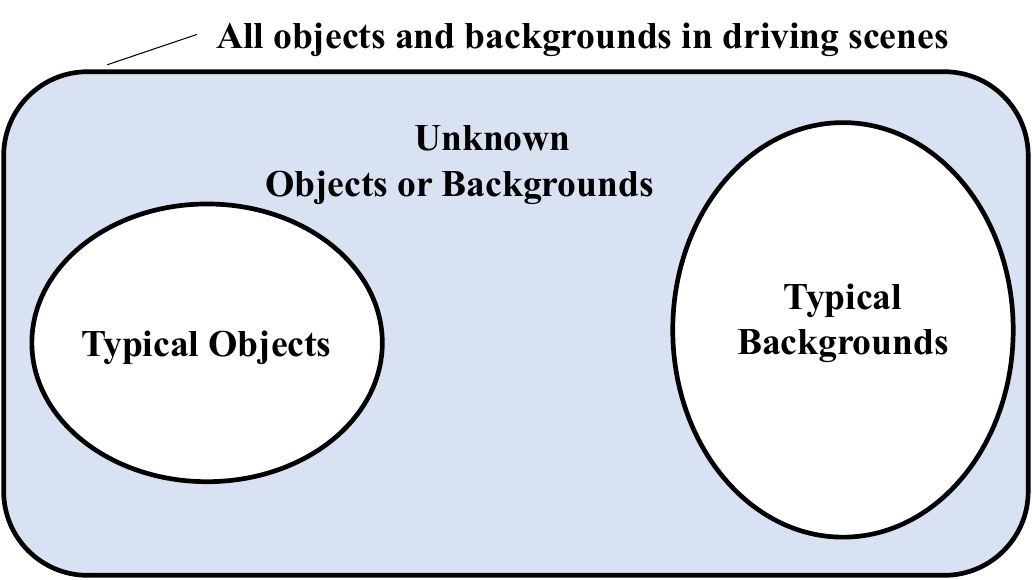}
\label{fig:venn_underlying}
}
\subfigure[]{
\includegraphics[width=4.cm]{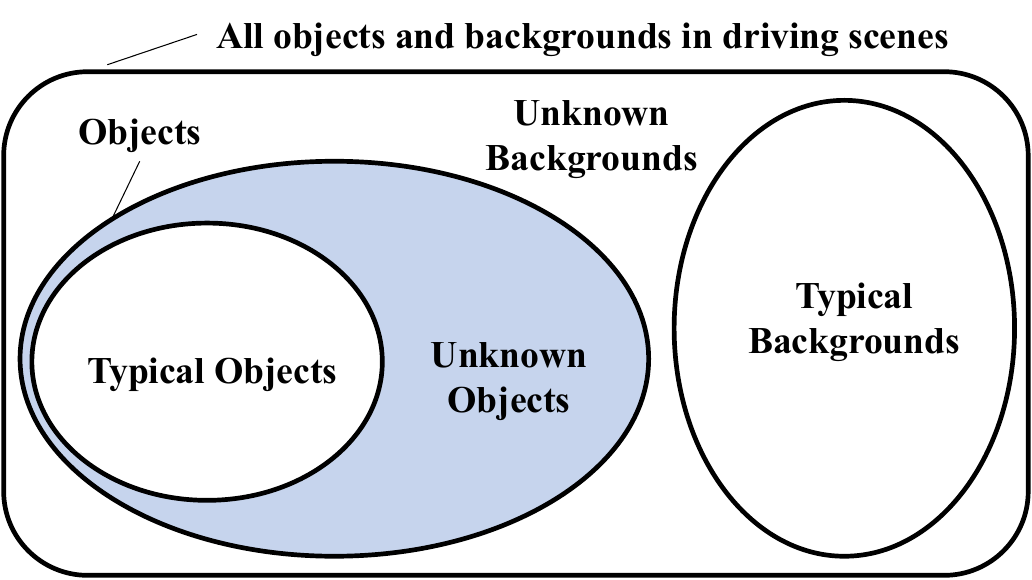}
\label{fig:venn_obj}
}
\caption{Venn diagrams comparing between the unknown score (Eg. \ref{eq:one_vs_rest_score}) and the unknown objectness score (Eg. \ref{eq:anomaly_score}). Typical objects and backgrounds indicate predefined semantic-segmentation classes, while unknown objects and backgrounds indicate ones that are not fall into the predefined classes. Road obstacles are classified into unknown objects.}
\label{fig:venn}
\end{figure}

The predicted probabilities obtained using the sigmoid head can be considered as a result of $K$ binary classifiers, where $K$ denotes the number of classes.
In this setting, an unknown pixel can be detected as a pixel with low predicted probabilities for all predefined classes.
Semantic-segmentation labels are usually assigned to nearly every pixel, thereby suggesting that the $K$ classes cover all foreground objects and backgrounds that could typically appear in images captured while driving.
Therefore, a pixel with low predicted probabilities for all predefined classes is atypical; that is, it represents an unknown pixel.
This is illustrated in the Venn diagram in Fig. \ref{fig:venn_underlying}.
Given the predicted probabilities of pixel $i$ as $p_{ik}$ $(k\in\{1,\dots,K\})$, the corresponding unknown score can be defined as
\begin{equation}
    S_i=\prod_{k=1}^K(1-p_{ik}).
    \label{eq:one_vs_rest_score}
\end{equation}
A well-trained segmentation model can assign a high probability, that nearly equals one, to pixels corresponding to a predefined class, while the corresponding unknown score nearly equals zero. On the other hand, the predicted pixel probabilities in an unknown region remain small with correspondingly large unknown scores.

\begin{figure}[t!]
\centering
\includegraphics[width=8.5cm]{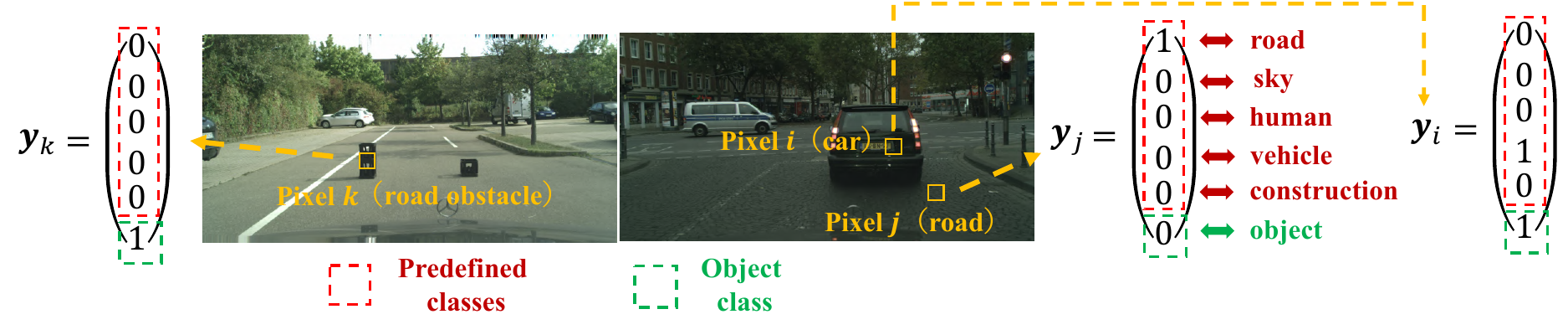}
\caption{Examples of labels for three typical types of pixels. This figure shows three pixels $i,j,k$ corresponding to a car, a road and a road obstacle pixels, respectively. Pixel $i$ is assigned to both the vehicle and object classes, while pixel $j$ is assigned to the road class only, because cars are objects, and roads are not. Pixel $k$ is assigned to the object class only, because road obstacles do not fall into any of the predefined classes, while they are objects.}
\label{fig:label_examples}
\end{figure}

\subsection{Unknown Objectness Scores}
\label{sec:prior_knowledge}

The unknown score defined in Eq. \ref{eq:one_vs_rest_score} can be used to detect pixels in unknown regions, not just unknown object pixels.
However, as with other pixel-wise anomaly-detection methods, this generic unknown scores tends to be unstable when applied to road obstacle detection.
Therefore, in this subsection, we extend the unknown score by incorporating the object detection fashions.
Because road obstacles are naturally classified as objects, this strategy can be expected to be effective to achieve stable performance.
We first define the pixel-wise objectness score $p_i^o$, which measures how much pixel $i$ is object-like.
Pixels of road obstacles are expected to possess high objectness scores.

To define $p_i^o$, we make an \textit{object class} by merging the foreground classes among classes predefined in semantic segmentation.
$p_i^o$ is defined by the predicted probability of the object class.
By adopting the sigmoid head, it is possible to assign labels of both a predefined class and the object class to a single pixel at the same time.
Figure \ref{fig:label_examples} illustrates how to make such labels more concretely.
Considering the objectness score $p_i^o$, the unknown score defined in Eq. \ref{eq:one_vs_rest_score} can be extended as follows.
\begin{equation}
    S_i=p^o_i\prod_{k=1}^K(1-p_{ik}).
    \label{eq:anomaly_score}
\end{equation}

\noindent
This score is referred to as the unknown objectness score.
As represented in Eq. \ref{eq:anomaly_score} and the Venn diagram in Fig. \ref{fig:venn_obj}, unknown objects can be detected as labels that are \textit{neither backgrounds nor typical objects, but are objects}.
As a result, we can focus on the more promising regions in an image where road obstacles could be located.
This facilitates a reduction in the number of false-positive predictions.

In addition, the proposed method can use OoD data optionally for supervision.
Specifically, if we obtain images that contain road obstacles and corresponding labels, we can use them for supervision by assigning the only object class to pixels corresponding to the road obstacles.
Figure \ref{fig:label_examples} shows label $\boldsymbol{y}_k$ of pixel $k$ corresponding to a road obstacle pixel. $\boldsymbol{y}_k$ has all 0 for elements of the predefined classes and 1 only for the element of object class.
In practical scenarios, it is common to have a limited number of road obstacle images, yet there is a desire to improve detection performance using these images. However, introducing a new class for road obstacles can result in a class imbalance issue. In the proposed method, road obstacle labels are used to emphasize the difference between typical and unknown objects (Fig. 3). This approach obviates the need for a new class and concurrently enhances detection performance.

\subsection{Boundary-aware Binary Cross-entropy Loss}
\label{sec:loss_function}
The flexibility of the sigmoid head could lead to an inherent disadvantage when evaluating the boundary regions in an image.
Whereas a segmentation classifier may confuse between at least two classes in boundary regions, a sigmoid classifier allows these regions to be assigned to two classes simultaneously.
This increases the unknown scores of these regions and results in false-positive predictions.
To overcome this disadvantage, we assign additional weights to the binary cross-entropy (BCE) loss incurred in the boundary regions.
The loss function $\mathcal{L}_n$ is expressed as
\begin{align}
    \mathcal{L}_n=
    -\frac{1}{N}\sum_{i=1}^N\sum_{k\in\mathcal{C}}
    f(y_{ik},p_{ik})
    -\frac{\lambda}{\sum_i\delta_i}\sum_{i=1}^N\delta_i\sum_{k\in\mathcal{C}}
    f(y_{ik},p_{ik}),
\label{eq:boundary_aware_loss}
\end{align}
\normalsize
where $f(y,p)=y\log p+(1-y)(1-\log p).$
In the above equation, the first term on the right of the equality indicates the standard BCE loss, and the second term indicates the additional boundary-aware loss.
$N$ denotes the number of pixels in image $n$, $y_{ik}$ denotes the label, and $\mathcal{C}$ denotes the set of classes that comprise the $K$ predefined classes and the object class. Moreover, $\delta_i=1$ when pixel $i$ lies in the boundary regions; in all other cases, $\delta_i=0$. The term $\lambda$ controls the weights of the boundary-aware loss.

\section{Experiments}
\label{sec:experiments}
\subsection{Experimental Setup}

\noindent
\textbf{Implementation Details.}\quad
In this study, the experiments were performed using DeepLabv3+ \cite{10.1007/978-3-030-01234-2_49} with the ResNet50 backbone \cite{7780459}.
The segmentation model was trained using the Cityscapes dataset \cite{Cordts2016Cityscapes}.
The optimizer used was SGD with a momentum value of 0.9 and weight decay of 0.0001. The initial learning rate was 0.01, and the ``poly'' learning rate policy was used with the power set to 0.9.
The value of $\lambda$ was set to three.

\noindent
\textbf{Datasets.}\quad
The performance of the proposed method was evaluated using three datasets comprising driving scene images that contain unknown road obstacles.
We first used the LostAndFound dataset \cite{10.1109/IROS.2016.7759186}, which is widely used and contains temporally sequential images with annotations of small unknown objects.
The test set used in this study comprises 1,203 images.
Next, we used the Fishyscapes Validation set \cite{Blum2021}, which comprises 100 images. This set is a subset of the LoadAndFound dataset and was created by selecting a suitable set of sequential images capturing a driving scene.
The images in this dataset contain additional annotations for pixels assigned to classes not considered in the Cityscapes dataset.
Furthermore, we used the road anomaly dataset \cite{9010663}, which comprises online images depicting a variety of unknown objects, such as animals and rocks.
This dataset contains 60 images depicting a variety of driving scenes, which are not limited solely to urban cityscapes.
The images in this dataset contain objects of various sizes and roads under various conditions. Identifying unknown objects from this dataset is, therefore, challenging.

\noindent
\textbf{OoD Data.}\quad
The state-of-the-art method proposed in \cite{Di_Biase_2021_CVPR} uses the ``void class'' in the Cityscapes dataset as OoD data for model training.
The ``void class'' is typically used to mask objects not assigned to any predefined class. Therefore, they can be considered pixels of unknown objects.
Following this procedure, the ``void class'' was used as the OoD data in the comparison between methods using OoD data. \cite{hendrycks2019oe,Di_Biase_2021_CVPR,10.1007/978-3-030-33676-9_3}.

\noindent
\textbf{Baselines.}\quad
Existing methods can be classified as those that use OoD data for supervision \cite{hendrycks2019oe,Di_Biase_2021_CVPR,10.1007/978-3-030-33676-9_3} and those that do not \cite{hendrycks17baseline,hendrycks2019anomalyseg,Ohgushi_2020_ACCV,Jung_2021_ICCV}.
To evaluate the performance of the methods proposed in \cite{Ohgushi_2020_ACCV,Di_Biase_2021_CVPR,Jung_2021_ICCV}, we used the source code implemented by the respective authors. The remaining methods were evaluated using the segmentation model and backbone identical to those considered in this study.

\noindent
\textbf{Evaluation Metrics.}\quad
We used three metrics widely adopted in existing research concerning obstacle detection studies.
The first is the area under the ROC curve (AUROC), which plots the true-positive rates against the false-positive rates at various threshold settings.
The second corresponds to average precision (AP), i.e., the area under the precision--recall curve.
The third metric corresponds to the false-positive rate at 95\% true-positive rate (FPR95).

\noindent
\textbf{Evaluation using the LostAndFound dataset.}\quad
The test images in LostAndFound contain not only road obstacles but also other unknown objects.
However, this dataset provides annotations for road obstacles exclusively. In other words, regardless of the identification of pixels corresponding to an unknown object, if those pixels do not indicate a road obstacle, they are considered false positives.
Therefore, additional processing is required to mask unknown objects other than road obstacles, thereby facilitating appropriate evaluation.
Following existing studies \cite{9010663,Ohgushi_2020_ACCV}, we utilized the ground-truth annotations to mask all pixels except those corresponding to the road and obstacles thereupon.

\begin{table*}[t!]
\caption{Performance comparison between proposed and existing methods.}
\centering
  \resizebox{0.95\textwidth}{!}{\begin{tabular}{c|c|ccc|ccc|ccc}
    \hline
    \multirow{2}{*}{Methods} & OoD & \multicolumn{3}{c|}{LostAndFound Test} & \multicolumn{3}{c|}{Fishyscapes Validation} & \multicolumn{3}{c}{Road Anomaly}  \\
     & Data &  FPR95$\downarrow$ & AP$\uparrow$ & AUROC$\uparrow$ & FPR95$\downarrow$ & AP$\uparrow$ & AUROC$\uparrow$ & FPR95$\downarrow$ & AP$\uparrow$ & AUROC$\uparrow$    \\
    \hline
  Softmax Entropy \cite{hendrycks17baseline}& & 19.45 & 39.58 & 95.40  & 30.55 & 7.92 & 91.38 & 70.93 & 17.06 & 69.35 \\
  Max Logit \cite{hendrycks2019anomalyseg}& & 16.44  & 53.06 & 96.91  & 30.74 & 12.59 & 93.89 & 68.03 & 18.64 & 72.82 \\
  Ohgushi et al. (2020) \cite{Ohgushi_2020_ACCV}& & 14.06 & 50.68 & 97.16 & 36.86 & 6.07 & 90.83 & 67.64 & 20.25 & 71.92 \\
  SML \cite{Jung_2021_ICCV}& & 35.51 & 39.65 & 92.87 & \textbf{14.53} & \textbf{36.56} & \textbf{96.34} & 50.91 & 25.32 & 81.37  \\
  Ours (w/o OoD data)& & \textbf{3.92} & \textbf{81.50} & \textbf{98.94} & 27.92 & 25.88 & 93.81 & \textbf{44.15} & \textbf{48.41} & \textbf{89.21}  \\
  \hline
  Outlier Exposure \cite{hendrycks2019oe}& \checkmark & 15.76 & 70.21 & 97.80 & 36.94 & 22.64 & 93.45 & 67.83 & 19.71 & 70.61 \\
  Outlier Head \cite{10.1007/978-3-030-33676-9_3}& \checkmark & 13.92 & 73.24 & 97.61 & 32.18 & 24.32 & 94.04 & 71.41 & 24.30 & 73.45 \\
  SynBoost \cite{Di_Biase_2021_CVPR}& \checkmark & 22.04 & 78.64 & 96.63  & 46.43 & 48.11 & 94.72 & 66.15 & 35.52 & 81.16 \\
  Ours (w/ OoD data)& \checkmark & \textbf{1.17} & \textbf{87.74} & \textbf{99.52} & \textbf{24.63} & \textbf{48.96} & \textbf{95.10} & \textbf{45.37} & \textbf{49.07} & \textbf{88.78}  \\
    \hline
  \end{tabular}}
  \label{table:main_result_fs_ra}
\end{table*}

\begin{table*}[t!]
\caption{Comparison between the results of the unknown scores (US) defined in Eq. \ref{eq:one_vs_rest_score} and the unknown objectness scores (UOS) defiend in Eq. \ref{eq:anomaly_score}.}
  \centering
  \resizebox{0.9\textwidth}{!}{\begin{tabular}{c|ccc|ccc|ccc}
    \hline
    \multirow{2}{*}{Methods} & \multicolumn{3}{c|}{LostAndFound Test} & \multicolumn{3}{c|}{Fishyscapes Validation} & \multicolumn{3}{c}{Road Anomaly}  \\
     &  FPR95$\downarrow$ & AP$\uparrow$ & AUROC$\uparrow$ & FPR95$\downarrow$ & AP$\uparrow$ & AUROC$\uparrow$ & FPR95$\downarrow$ & AP$\uparrow$ & AUROC$\uparrow$    \\
    \hline
  US (w/o OoD data) & 29.49 & 33.96 & 93.83 & 43.92 & \textbf{32.19} & 91.53 & 72.73 & 23.26 & 70.64  \\
  UOS (w/o OoD data) & \textbf{3.92} & \textbf{81.50} & \textbf{98.94} & \textbf{27.92} & 25.88 & \textbf{93.81} & \textbf{44.15} & \textbf{48.41} & \textbf{89.21}  \\
  \hline
  US (w/ OoD data)& 18.18 & 69.86 & 96.99  & 37.25 & \textbf{49.38} & 93.94 & 68.88 & 33.74 & 77.88 \\
  UOS (w/ OoD data)& \textbf{1.17} & \textbf{87.74} & \textbf{99.52} & \textbf{24.63} & 48.96 & \textbf{95.10} & \textbf{45.37} & \textbf{49.07} & \textbf{88.78}  \\
    \hline
  \end{tabular}}
\label{table:comparsion_uas_so}
\end{table*}

\subsection{Evaluation Results}
\label{sec: results}
Table \ref{table:main_result_fs_ra} presents the experimental results obtained by applying the proposed method to the LostAndFound Test, Fishyscapes Validation, and Road Anomaly datasets.
The performances of the candidate methods are first compared considering the setting without OoD data.
Among the five methods presented in the top row of Table \ref{table:main_result_fs_ra}, the proposed method outperforms the other approaches when applied to the LostAndFound Test and Road Anomaly datasets.
SML \cite{Jung_2021_ICCV} demonstrates better performance than our method on the Fishyscapes Validation dataset, but its performance is inferior to that of our method by a large margin on the LostAndFound Test set.
The performance of the methods presented at the bottom of Table \ref{table:main_result_fs_ra} were compared under the OoD data setting.
The proposed method demonstrates the best performance for all the datasets.

\subsection{Effectiveness of Objectness Scores}
\label{sec:effectiveness_of_objectnessscore}
Table \ref{table:comparsion_uas_so} shows the comparison between the results of the unknown scores defined in Eq. \ref{eq:one_vs_rest_score} and those of the unknown objectness scores defined in Eq. \ref{eq:anomaly_score}. We can see that most of the evaluation metrics are improved by considering the objectness scores, both with and without using OoD data. In particular, FPR95 values are significantly reduced due to the reduction of false positive predictions in the background regions. On the other hand, when pixels corresponding to road obstacles have low objectness scores, it can lead to reduced true positive predictions. In fact, this is the reason why AP values of the unknown objectness scores are slightly lower than those of the unknown scores for the evaluation using Fishyscapes Validation.

\subsection{Grouping of Original Classes}
Although labelled data with 19 classes are generally used in the Cityscapes dataset,
these classes can be grouped into seven large categories---``road," ``flat(w/o road)," ``human," ``vehicle," ``construction," ``object," and ``background" \cite{Cordts2016Cityscapes}.
Table \ref{table:group_class} presents the performance of the proposed approach when applied to the Fishyscapes Validation set between considering the grouped 7 and the original 19 classes under the with and without OoD data settings. As can be seen, the proposed method demonstrates better performance when trained using 7 rather than 19 classes.
This is because some of the 19 classes are rare classes that do not appear frequently in the training data.
Pixels corresponding to such classes are often ambiguously predicted under other similar classes.
This indicates that such pixels are more likely to achieve higher unknown scores and yield false-positive results.
Therefore, incorporating rare classes under a large category can facilitate a reduction in the prediction ambiguity among similar classes and improve performance.

\begin{table}[t]
  \footnotesize
  \caption{Comparison between training results obtained considering the original 19 Cityscapes and 7 grouped classes.}
  \centering
  \begin{tabular}{c|c|ccc}
    \hline
    \multirow{2}{*}{Methods} & OoD & \multicolumn{3}{c}{Fishyscapes Validation} \\
     & data & FPR95 $\downarrow$ & AP $\uparrow$ & AUROC $\uparrow$ \\
    \hline
  Ours w/ 19 classes & & 27.92 & 25.88 & 93.81    \\
  Ours w/ 7 classes &  & \textbf{25.49}  & \textbf{32.43} & \textbf{93.82} \\
  \hline
  Ours w/ 19 classes & \checkmark & 24.63 & 48.96 & 95.10 \\
  Ours w/ 7 classes & \checkmark & \textbf{9.85} & \textbf{66.11} & \textbf{98.26} \\
    \hline
  \end{tabular}
  \label{table:group_class}
\end{table}

\begin{table}[t]
\footnotesize
\caption{Comparison of Cityscapes mIoU values.}
\centering
\begin{tabular}{c|c|c}
\hline
     Methods & OoD data & Cityscapes mIoU  \\
\hline
     Softmax Entropy \cite{hendrycks17baseline} &  & 77.74   \\
     Outlier Exposure \cite{hendrycks2019oe} & \checkmark & 68.83  \\
     Outlier Head \cite{10.1007/978-3-030-33676-9_3} & \checkmark & 77.27  \\
     Ours & \checkmark &  76.85  \\
\hline
\end{tabular}
\label{table:semantic_segmentation_performance}
\end{table}

\begin{table}[t]
\footnotesize
\caption{Comparison between computational times per image required by different candidate models measured over 100 repetitions of the experiment.}
\centering
\begin{tabular}{c|c}
\hline
     Methods & Computational time per image (ms)  \\
\hline
     SynBoost \cite{Di_Biase_2021_CVPR} & 190.97 $\pm$ 18.37  \\
     Ohgushi et al. (2020) \cite{Ohgushi_2020_ACCV} & 125.10 $\pm$ 4.39 \\
     SML \cite{Jung_2021_ICCV} & 20.79 $\pm$ 2.20  \\
     Ours & 15.57 $\pm$ 2.19  \\
\hline
\end{tabular}
\label{table: computational_cost}
\end{table}

\begin{figure*}[t]
\centering
\includegraphics[width=17.8cm]{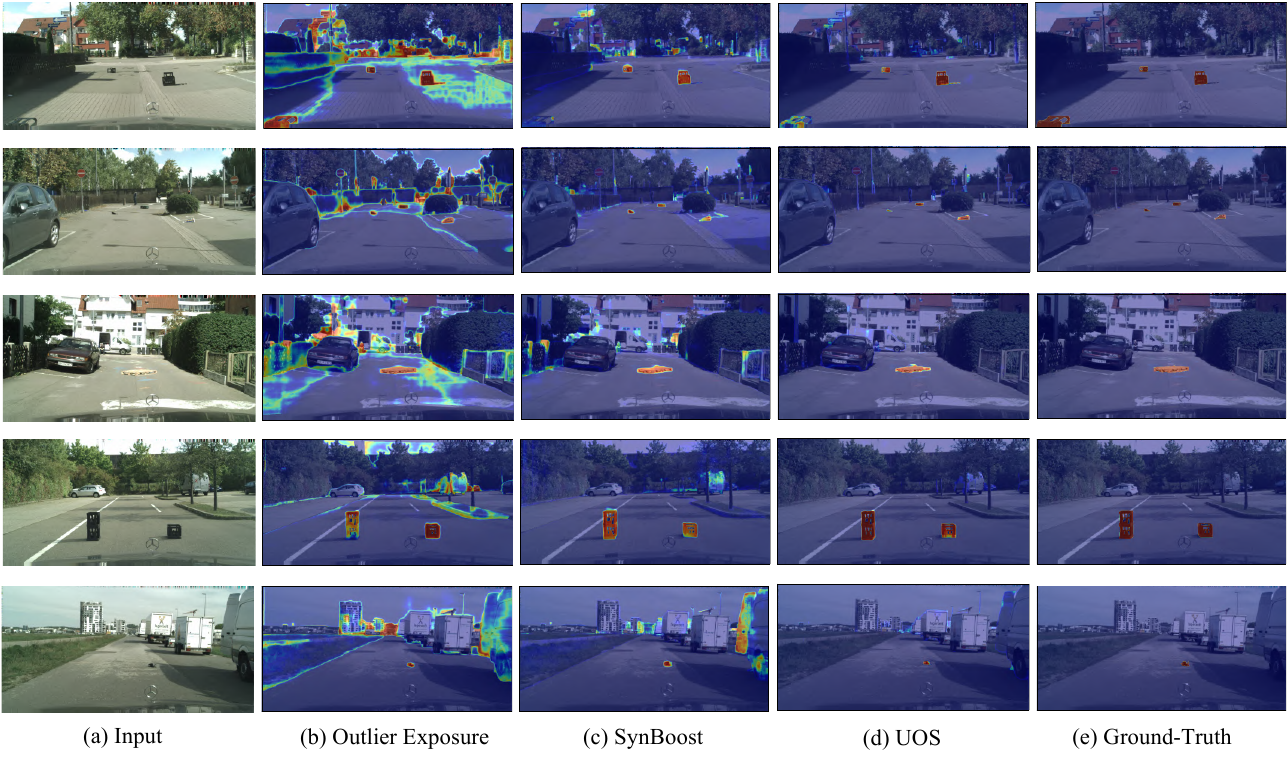}
\caption{Qualitative results for Ourtlier Exposure, Synboost, and the unknown objectness score (UOS).}
\label{fig:qual_results}
\end{figure*}

\subsection{Comparison to Softmax Heads}
When segmentation models are trained with a softmax head, uncertainty scores, such as entropy \cite{hendrycks17baseline} or max logit \cite{hendrycks2019anomalyseg} can be used instead of the unknown scores defined in Eq. \ref{eq:one_vs_rest_score}.
However, when trained with OoD data, the softmax head is possible to degrade the segmentation performance.
Outlier Exposure \cite{hendrycks2019oe} is a technique to yield low maximum softmax values for OoD data by minimizing the cross-entropy loss between model predictions and uniform distribution.
As shown in Tab. \ref{table:semantic_segmentation_performance}, the mIoU of Outlier Exposure is lower than the others. 
The predictions performed using a softmax head are based on the assumption that each pixel is assigned to exactly one class. That is, it is not suitable for representing ambiguous states.
On the other hand, predictions performed using a sigmoid head can be more suitable, because the sigmoid head allows each pixel to be assigned to multiple classes simultaneously or not to any classes.
In fact, we can see that the performance degradation of the proposed method is mitigated compared to Outlier Exposure.

\subsection{Comparison of Computational Costs}
Table \ref{table: computational_cost} compares the computational time per image required by the state-of-the-art methods \cite{Ohgushi_2020_ACCV,Jung_2021_ICCV,Di_Biase_2021_CVPR}.
All experiments were performed on a workstation equipped with Tesla V100 GPUs.
To facilitate a fair comparison, the computational times were measured using the same segmentation model as ours.
The methods based on the perceptual difference \cite{Ohgushi_2020_ACCV,Di_Biase_2021_CVPR} require the use of extra networks, hence, incurring additional computational time compared to other approaches.
The method proposed in \cite{Jung_2021_ICCV} also does not require any extra networks, and thus its computational time is close to the proposed method. However, its computational time is longer than the proposed method due to the additional post-processing required.

\subsection{Qualitative Evaluation}

Figure 5 presents the qualitative results for Outlier Exposure, SynBoost, and the unknown objectness score. It is evident that the use of the unknown object score significantly reduces the number of false negative samples compared to the other two methods. This reduction can be attributed to the fact that the unknown objectness score explicitly considers objectness, which substantially lowers the anomaly scores in the background regions.

\section{Limitations}
The performance of the proposed method depends on the objectness scores. This means that the method's ability to detect road obstacles may be compromised if it cannot recognize the pixels of these obstacles as objects. Therefore, devising strategies to grasp more universal object concepts stands as a crucial direction for future research.

\section{Conclusions}
In this paper, we presented a novel method for identifying unknown objects in images of driving scenarios.
The method is based on a semantic-segmentation network with a sigmoid head that allows us to obtain objectness scores and predicted probabilities of predefined classes simultaneously. 
By incorporating the objectness score into the unknown scores, our method is capable of stably detecting unknown objects in driving scenes. 
We demonstrated the effectiveness of our approach by evaluating it on publicly available datasets and showing that it outperforms state-of-the-art methods. 
Our results indicate that the proposed method has the potential to enhance the safety of autonomous driving systems by enabling the detection of unknown road obstacles.

\clearpage

\bibliographystyle{IEEEtran}
\bibliography{IEEEexample}


\end{document}